\documentclass[pdflatex,sn-nature]{sn-jnl}

\usepackage{graphicx}%
\usepackage{multirow}%
\usepackage{amsmath,amssymb,amsfonts}%
\usepackage{amsthm}%
\usepackage{mathrsfs}%
\usepackage[title]{appendix}%
\usepackage{xcolor}%
\usepackage{textcomp}%
\usepackage{manyfoot}%
\usepackage{booktabs}%
\usepackage{algorithm}%
\usepackage{algorithmicx}%
\usepackage{algpseudocode}%
\usepackage{listings}%
\usepackage{placeins}

\usepackage{tikz}
\usepackage{tabularx}
\usepackage{array}
\usepackage{threeparttable}
\usepackage{adjustbox}
\usepackage{longtable}

\usepackage{pgfplotstable}
\pgfplotsset{compat=1.18}

\usetikzlibrary{arrows.meta,positioning,shapes.geometric,calc}

\begin{document}

\title[Agentic induction of common data elements and information extraction from sarcoma case report abstracts]{Agentic induction of common data elements and information extraction from sarcoma case report abstracts}


\author*[1,2]{\fnm{Daniel} \sur{Reichenpfader}}\email{daniel.reichenpfader@bfh.ch}

\author[3]{\fnm{Kwaku} \sur{Appiagyei}}

\author[4]{\fnm{Manovah} \sur{Nadarajah}}

\author[5]{\fnm{Ahmed Reda} \sur{Bahr}}

\author[6]{\fnm{Irina} \sur{Filchenko}}

\author[7,8,9]{\fnm{Fabio} \sur{Dennstädt}}

\affil*[1]{\orgdiv{Institute for Patient-Centered Digital Health}, \orgname{Bern University of Applied Sciences}, \orgaddress{\city{Bern}, \country{Switzerland}}}

\affil[2]{\orgdiv{Faculty of Medicine}, \orgname{University of Geneva}, \orgaddress{\city{Geneva}, \country{Switzerland}}}

\affil[3]{\orgname{F. Hoffmann-La Roche AG}, \orgaddress{\city{Basel}, \country{Switzerland}}}

\affil[4]{\orgname{Kantonsspital Aarau}, \orgaddress{\city{Aarau}, \country{Switzerland}}}

\affil[5]{\orgname{University of Alexandria}, \orgaddress{\city{Alexandria}, \country{Egypt}}}

\affil[6]{\orgdiv{Department of Neurology}, \orgname{Inselspital, Bern University Hospital and University of Bern}, \orgaddress{\city{Bern}, \country{Switzerland}}}

\affil[7]{\orgname{Data Semantics and Decision Making Institute (DSDMI)}, \orgaddress{\city{St.\ Gallen}, \country{Switzerland}}}

\affil[8]{\orgdiv{Department of Radiation Oncology}, \orgname{Inselspital, Bern University Hospital and University of Bern}, \orgaddress{\city{Bern}, \country{Switzerland}}}

\affil[9]{\orgdiv{Department of Digital Medicine}, \orgname{Inselspital, Bern University Hospital and University of Bern}, \orgaddress{\city{Bern}, \country{Switzerland}}}

\abstract{Common data elements (CDEs) support clinical data comparability but are usually defined manually. We investigated whether large language model agents could derive a CDE schema from sarcoma case report abstracts and apply it at scale. An agent proposed 72,909 candidate CDEs from 24,588 PubMed abstracts; agentic consolidation yielded 1,570 CDEs. We retained 53 supported by at least 50 abstracts and attempted evidence-linked extraction across the corpus. Processing succeeded for all retained CDEs in 79.2\% of abstracts, including responses indicating non-mention. Two human reviewers approved 83.0\% of retained CDEs for relevance, 67.9\% for clarity, and 56.6\% for correctness. In 100 sampled abstracts, exact agreement with the value extracted by the LLM was 94.9\% among 1,116 pairs with a non-empty answer (abstract-level bootstrap 95\% CI, 93.0--96.4\%). Errors concentrated in anatomical and temporal fields. These findings support agent-assisted schema drafting followed by expert curation before large-scale extraction.}


\maketitle


\section{Introduction}
Sarcomas are a rare and heterogeneous group of malignancies differing in histology, anatomic origin, molecular alterations, treatment, and clinical course \cite{sbaraglia2019pathology}. They account for about 5.6 new cases per 100,000 persons per year in Europe \cite{stillerDescriptiveEpidemiologySarcomas2013}. A case report is a detailed description of the presentation, diagnosis, treatment, and outcome of an individual patient \cite{nissenClinicalCaseReport2014}. Case reports and case series contribute to the recognition of new diseases and the course of rare medical scenarios \cite{vandenbrouckeDefenseCaseReports2001}. Their value as a collective resource is limited by narrative presentation, inconsistent terminology, and uneven reporting. The CAse REport (CARE) guidelines support more complete and transparent case reporting, but inconsistent and unstandardised reporting remains common in case reports \cite{gagnier2014care, yangAdherenceUnpublishedCase2022}.

Common data elements (CDEs) were introduced within the US National Institutes of Health (NIH) to standardise data collection in clinical trials \cite{nihCDErepository}. Each CDE combines a clearly defined concept or question with a specified response format, including the data type, permissible values, and units where applicable, enabling consistent recording and reuse across studies \cite{sheehan2016cde}. Beyond clinical trials, CDEs are used for real-world applications, because they define directly usable data \cite{huser2018common}. Examples include the oncology-specific mCODE standard \cite{ostermanImprovingCancerData2020} and LLM-assisted mapping of local variables onto NIH CDEs \cite{wangCDEMapperEnhancingNational2025}.


In conventional CDE definition workflows, domain experts usually define a set of CDEs, also known as a CDE schema, before reviewing the entire input corpus. This provides control over the data elements and their meaning, but requires substantial domain understanding and can overlook concepts that emerge only when a diverse corpus is examined. As a result, a (semi-)automated system for deriving CDE schemas directly from an input corpus would be very valuable. Outside the CDE context, large language models (LLMs) are being used to populate a user-supplied knowledge schema \cite{caufieldStructuredPromptInterrogation2024}, to learn ontological structure \cite{babaeigiglouLLMs4OLLargeLanguage2023}, or to induce and canonicalise a schema from text for knowledge-graph construction \cite{zhangExtractDefineCanonicalize2024,baiAutoSchemaKGAutonomousKnowledge2026}. For CDEs specifically, embeddings and clustering have been used to group existing elements \cite{krishnamurthyDynamicFrameworkSemantic2025}, but not to derive new ones from a literature corpus.

\begin{figure}[htbp]
\centering
\includegraphics[width=\textwidth]{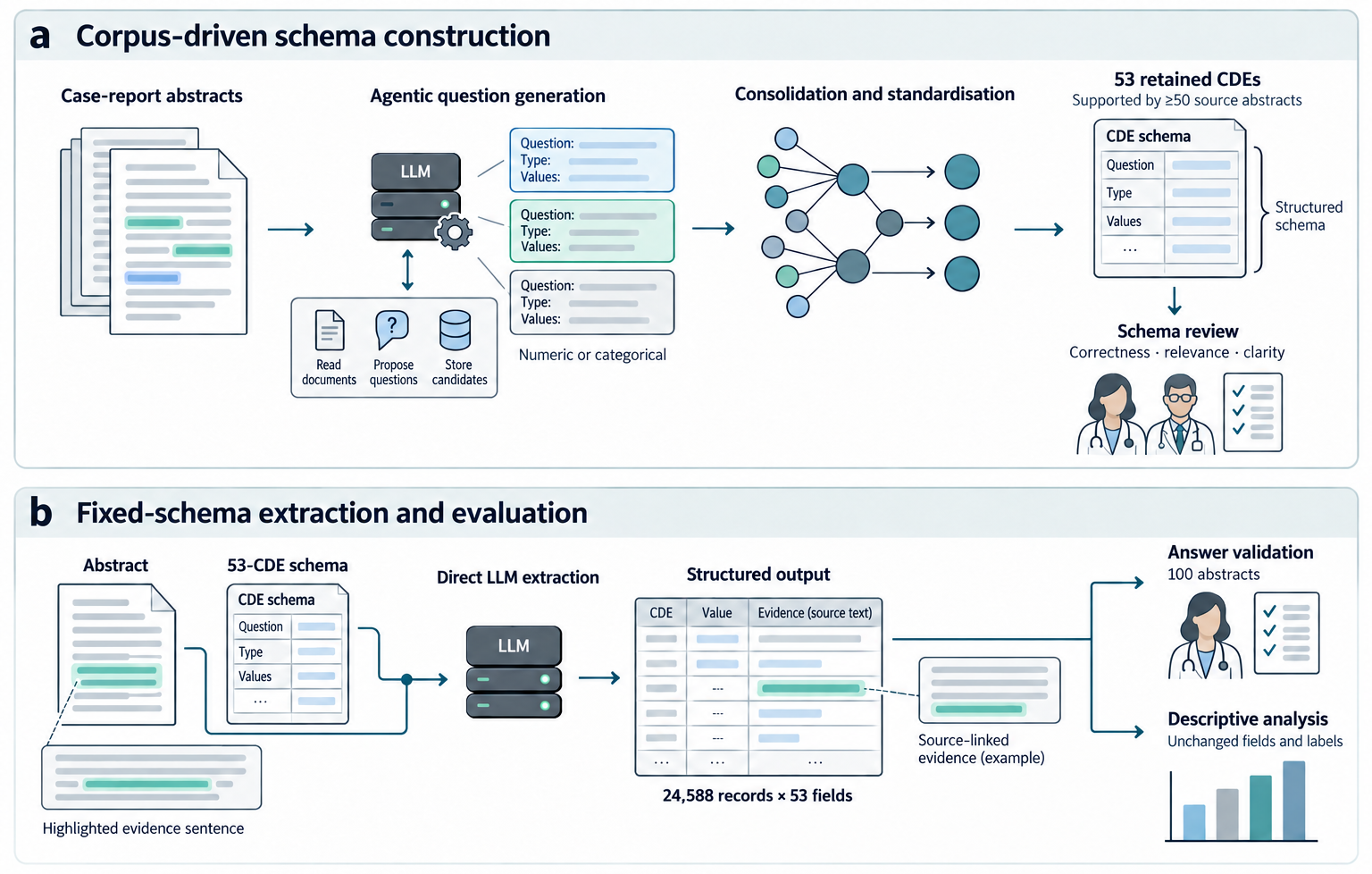}
\caption{\textbf{Illustrative overview of the study workflow.} (a) An LLM agent proposed CDEs for each sarcoma case report abstract in an input corpus. Next, these CDEs were consolidated and standardised, resulting in an induced CDE schema. We selected 53 CDEs supported by at least 50 source abstracts. Two domain experts assessed schema correctness, relevance, and clarity. (b) Values for all 53 CDEs were extracted from each abstract using standard prompting. Every extracted answer was linked to an evidence text span in the source abstract. We validated the LLM-derived outputs using a subset of 100 abstracts. Furthermore, we performed a descriptive analysis based on the whole corpus of 24,577 unique abstracts.}
\label{fig:pipeline}
\end{figure}

LLMs were successfully applied to extracting predefined data elements from biomedical publications \cite{dennstaedt2026trialclassification,windisch2024metastatic} and clinical text \cite{agrawalLargeLanguageModels2022a, dengComorbidityClassificationClinical2026}, a task commonly referred to as (biomedical) information extraction (IE) \cite{Ed3book_jan26}. For example, Zhang et al. investigated LLM-based extraction of clinical information from case reports on inborn errors of metabolism and introduced CaseReportBench, an expert-annotated benchmark comprising 138 reports with information organised into 14 predefined clinical categories \cite{zhangCaseReportBenchLLMBenchmark2025a}. In our earlier study, an LLM classified radiotherapy case report abstracts using 23 predefined boolean CDEs with the model gpt-oss-120b achieving an overall F1 score of 93.64\% \cite{dennstaedt2025classification}. While these studies demonstrate successful data extraction using LLMs based on investigator-specified data schemas, they do not establish whether an LLM can also construct a useful CDE schema from an input corpus of biomedical literature. This study aimed to assess whether LLM agents can derive a clinically useful CDE schema from sarcoma case report abstracts and whether the resulting schema can support accurate information extraction at scale. An overview of the study workflow is shown in \autoref{fig:pipeline}.

\section{Results}
\label{sec:results}

\subsection{Corpus construction and CDE schema induction}

The PubMed search for sarcoma case reports returned 58,265 records, of which 42,018 (72.1\%) had an abstract. The LLM-based classifier retained 24,588 abstracts as single-patient case reports, corresponding to 58.5\% of records with abstracts and 42.2\% of all retrieved records. Abstracts retrieved at this stage were used for per-abstract CDE proposal. Subsequently, eleven duplicate abstracts were removed, leaving 24,577 unique abstracts for descriptive analysis (Figure ~\ref{fig:flow}A)).

\begin{figure}[htbp]
\centering

\begin{minipage}[t]{0.48\textwidth}
\vspace{0pt}
\raggedright
{\small\bfseries (A) Corpus construction}
\par\medskip
\centering

\resizebox{\linewidth}{!}{%
\begin{tikzpicture}[
  node distance=8mm,
  box/.style={
    rectangle,
    rounded corners=2pt,
    draw=black,
    line width=0.6pt,
    fill=gray!7,
    align=center,
    text width=3.5cm,
    minimum height=1.05cm,
    inner sep=4pt,
    font=\small
  },
  side/.style={
    rectangle,
    rounded corners=2pt,
    draw=black!60,
    line width=0.5pt,
    fill=white,
    align=center,
    text width=2.3cm,
    minimum height=1.05cm,
    inner sep=4pt,
    font=\small
  },
  arrow/.style={
    -{Latex[length=2mm]},
    line width=0.7pt
  }
]

\node[box] (retrieved) {
  PubMed records retrieved\\
  \textbf{58,265}
};

\node[box, below=of retrieved] (abstracts) {
  Records with abstracts\\
  \textbf{42,018}
};

\node[box, below=of abstracts] (eligible) {
  Classified as single-patient\\
  case reports\\
  \textbf{24,588}
};

\node[box, below=of eligible] (unique) {
  Unique abstracts for\\
  descriptive analysis\\
  \textbf{24,577}
};

\node[side, right=4mm of retrieved] (missing) {
  No abstract\\
  \textbf{16,247}
};

\node[side, right=4mm of abstracts] (excluded) {
  Excluded by classifier\\
  \textbf{17,430}
};

\node[side, right=4mm of eligible] (duplicates) {
  Duplicate abstracts\\
  \textbf{11}
};

\draw[arrow] (retrieved) -- (abstracts);
\draw[arrow] (abstracts) -- (eligible);
\draw[arrow] (eligible) -- (unique);

\draw[arrow] (retrieved.east) -- (missing.west);
\draw[arrow] (abstracts.east) -- (excluded.west);
\draw[arrow] (eligible.east) -- (duplicates.west);

\end{tikzpicture}%
}
\end{minipage}
\hfill
\begin{minipage}[t]{0.48\textwidth}
\vspace{0pt}
\raggedright
{\small\bfseries (B) CDE schema induction}
\par\medskip
\centering

\resizebox{\linewidth}{!}{%
\begin{tikzpicture}[
  node distance=5mm,
  stage/.style={
    rectangle,
    rounded corners=2pt,
    draw=black,
    line width=0.6pt,
    fill=gray!7,
    align=center,
    text width=6.2cm,
    minimum height=0.95cm,
    inner sep=5pt,
    font=\small
  },
  arrow/.style={
    -{Latex[length=2mm]},
    line width=0.7pt
  }
]

\node[stage] (proposed) {
  \textbf{Per-abstract CDE proposal}\\
  72,909 proposed CDEs from 24,588 abstracts
};

\node[stage, below=of proposed] (deduplicated) {
  \textbf{Exact question--type deduplication}\\
  7,788 unique CDEs
};

\node[stage, below=of deduplicated] (passone) {
  \textbf{First agentic consolidation pass}\\
  2,183 CDEs
};

\node[stage, below=of passone] (passtwo) {
  \textbf{Second agentic consolidation pass}\\
  1,570 CDEs: 560 numeric, 1,010 categorical
};

\node[stage, below=of passtwo] (standardised) {
  \textbf{Categorical value-set standardisation}\\
  626 of 1,010 vocabularies revised (62.0\%)\\
  63 synonym relationships recorded in 39 CDEs
};

\node[stage, below=of standardised,
      fill=gray!15, line width=0.9pt] (retained) {
  \textbf{Selection: support from $\geq 50$ abstracts}\\
  53 retained CDEs: 14 numeric, 39 categorical
};

\draw[arrow] (proposed) -- (deduplicated);
\draw[arrow] (deduplicated) -- (passone);
\draw[arrow] (passone) -- (passtwo);
\draw[arrow] (passtwo) -- (standardised);
\draw[arrow] (standardised) -- (retained);

\end{tikzpicture}%
}

\par\medskip
{\footnotesize
\textbf{Support among the 1,570 consolidated CDEs}
\par\smallskip
\begin{tabular}{@{}lrr@{}}
\toprule
Source abstracts & CDEs & \% \\
\midrule
Exactly 1 & 1,092 & 69.6 \\
$\geq 2$  & 478   & 30.4 \\
$\geq 10$ & 154   & 9.8 \\
$\geq 50$ & 53    & 3.4 \\
\bottomrule
\end{tabular}
\par\smallskip
Support counts distinct source abstracts;
the $\geq$ thresholds are cumulative.
}
\end{minipage}

\caption{\textbf{Corpus construction and CDE schema induction.}
(A) Retrieval, screening, and removal of duplicate abstracts.
(B) Proposal, deduplication, consolidation, standardisation,
and selection of CDEs.
Schema induction and information extraction used all 24,588
accepted abstracts; duplicate removal applied only to the
descriptive analysis.
Exact deduplication retained numeric and categorical CDEs
separately, including 30 question texts occurring with both types.
Value-set standardisation revised categorical vocabularies
without reducing the number of CDEs.}
\label{fig:flow}
\end{figure}

The workflow reduced 72,909 proposed CDEs to 1,570 consolidated elements. Of these, 53 (14 numeric and 39 categorical) were supported by at least 50 distinct source abstracts and retained for corpus-wide information extraction (Figure~\ref{fig:flow}B).


\begin{table}[htbp]
\centering
\caption{Reviewer approval and agreement for the 53 retained CDEs.}
\label{tab:schema-review}
\small
\begin{tabularx}{\textwidth}{@{}Xrrrrr@{}}
\toprule
Dimension &
\shortstack{Reviewer 1\\approval} &
\shortstack{Reviewer 2\\approval} &
\shortstack{Reviewer\\agreement} &
$\kappa$ &
\shortstack{Approval by\\both reviewers} \\
\midrule
Correctness & 48/53 & 30/53 & 35/53 (66.0\%) & 0.24 & 30/53 (56.6\%) \\
Relevance   & 48/53 & 47/53 & 46/53 (86.8\%) & 0.29 & 44/53 (83.0\%) \\
Clarity     & 39/53 & 50/53 & 36/53 (67.9\%) & -0.10 & 36/53 (67.9\%) \\
\bottomrule
\end{tabularx}
\par\smallskip
{\footnotesize
All 53 CDEs had complete ratings.
Agreement includes matching approvals and non-approvals;
approval by both reviewers counts only matching approvals.
$\kappa$: Cohen's Kappa.
}
\end{table}
Of the 53 CDEs, 30 (56.6\%) were approved by both reviewers for correctness, 44 (83.0\%) for relevance, and 36 (67.9\%) for clarity. Nineteen CDEs (35.8\%) were approved by both reviewers on all three dimensions. Inter-reviewer agreement, counting both concordant approval and concordant non-approval, was 66.0\% for correctness ($\kappa$ = 0.24), 86.8\% for relevance ($\kappa$ = 0.29), and 67.9\% for clarity ($\kappa$ = $-$0.10). Reviewer-specific approval counts and agreement statistics are summarised in \autoref{tab:schema-review}.

\subsection{CDE extraction and evaluation}

Of the 24,588 abstracts, 19,482 (79.2\%) were successfully processed for all 53 CDEs, including valid responses indicating non-mention. For the remaining 5,106 (20.8\%), at least one CDE could not be filled, affecting 6,287 of 1,303,164 abstract--CDE pairs (0.5\%). Error analysis showed that failure was most often due to an evidence span that did not occur in the source abstract (60.9\%), followed by proposed labels outside the value set (24.3\%), labels rejected as non-atomic (8.8\%), and missing evidence (5.9\%).


The evaluation sample comprised 5,300 abstract--CDE pairs (100 randomly chosen abstracts $\times$ 53 CDEs). Both the pipeline and reference were empty in 4,184 pairs; in the remaining 1,116 pairs, at least one contained an answer (mean 11.16 evaluable CDEs per abstract). The pipeline exactly matched the reference in 1,059 of these 1,116 pairs (94.9\%; abstract-level bootstrap 95\% CI 93.0--96.4\%). Of the 57 non-matches, 34 (59.6\%) were pipeline-only answers, 22 (38.6\%) were different non-empty answers, and one (1.8\%) was a reference answer missed by the pipeline.
Field-specific comparison counts ranged from 0 to 77, including one comparison for total radiation dose and two for histological subtype (Supplementary Table~S1).

Answer-level exact agreement was high for commonly populated demographic, diagnostic, treatment, and outcome variables in the evaluation sample (selected results in Table~\ref{tab:accuracy}). Errors were more frequent for anatomical locations and temporal concepts: primary lesion location was correct in 27 of 39 cases (69.2\%), metastatic tumour location in 18 of 24 (75.0\%), granulocytic-sarcoma anatomical sites in 11 of 16 (68.8\%), and latency to a secondary tumour in 7 of 10 (70.0\%). Detailed results are provided in Supplementary Table~S1.

\begin{table}[htbp]
\centering
\caption{Selected results for answer-level exact agreement.}
\label{tab:accuracy}
\begin{tabularx}{\textwidth}{@{}>{\raggedright\arraybackslash}Xrrrr@{}}
\toprule
CDE & Matches & Non-matches & Compared & Agreement (\%) \\
\midrule
All CDEs & 1,059 & 57 & 1,116 & 94.9 \\
Patient sex & 75 & 2 & 77 & 97.4 \\
Age at diagnosis & 73 & 1 & 74 & 98.6 \\
Histopathologic diagnosis (biopsy) & 50 & 0 & 50 & 100.0 \\
Distant metastases & 44 & 2 & 46 & 95.7 \\
Tumour recurrence during follow-up & 37 & 0 & 37 & 100.0 \\
Surgical resection performed & 54 & 3 & 57 & 94.7 \\
Primary lesion location & 27 & 12 & 39 & 69.2 \\
Metastatic tumour location & 18 & 6 & 24 & 75.0 \\
Granulocytic-sarcoma anatomical sites & 11 & 5 & 16 & 68.8 \\
Latency to secondary tumour & 7 & 3 & 10 & 70.0 \\
\bottomrule
\end{tabularx}
\par\smallskip
{\footnotesize Compared pairs have a non-empty answer in the pipeline output, the reference, or both. Jointly empty pairs are excluded.}
\end{table}

\subsection{Coverage and descriptive analysis of the extracted corpus}

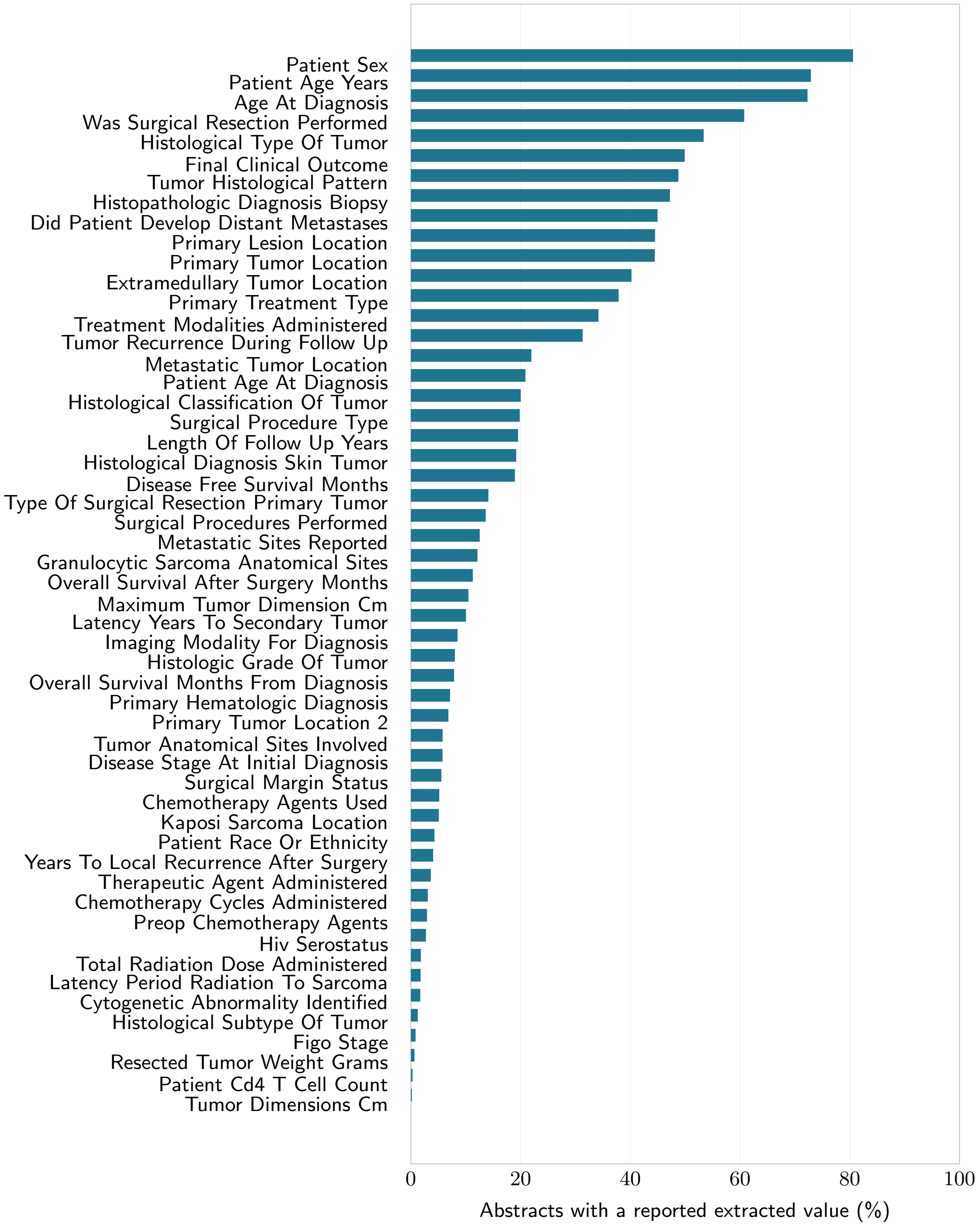
\begin{figure}[htbp]
\centering
\begin{tikzpicture}

\definecolor{coverageDark}{HTML}{207590}

\pgfplotstableread[col sep=comma]{
id,filled
1,19806
2,17917
3,17770
4,14933
5,13117
6,12266
7,11985
8,11604
9,11050
10,10938
11,10926
12,9884
13,9306
14,8402
15,7696
16,5397
17,5132
18,4919
19,4877
20,4801
21,4717
22,4660
23,3471
24,3349
25,3084
26,2982
27,2774
28,2578
29,2467
30,2090
31,1969
32,1935
33,1755
34,1680
35,1421
36,1417
37,1365
38,1271
39,1249
40,1057
41,996
42,892
43,757
44,721
45,675
46,448
47,434
48,419
49,308
50,210
51,159
52,73
53,43
}\coveragedata

\begin{axis}[
    width=0.64\textwidth,
    height=0.91\textheight,
    scale only axis,
    font=\sffamily\scriptsize,
    title style={
        font=\sffamily\bfseries\small,
        at={(0,1)},
        anchor=south west,
        align=left,
        yshift=2mm
    },
    xbar,
    bar width=5.5pt,
    xmin=0,
    xmax=100,
    ymin=-2,
    ymax=56,
    y dir=reverse,
    xtick={0,20,40,60,80,100},
    ytick={1,...,53},
    yticklabels={
        Patient Sex,
        Patient Age Years,
        Age At Diagnosis,
        Was Surgical Resection Performed,
        Histological Type Of Tumor,
        Final Clinical Outcome,
        Tumor Histological Pattern,
        Histopathologic Diagnosis Biopsy,
        Did Patient Develop Distant Metastases,
        Primary Lesion Location,
        Primary Tumor Location,
        Extramedullary Tumor Location,
        Primary Treatment Type,
        Treatment Modalities Administered,
        Tumor Recurrence During Follow Up,
        Metastatic Tumor Location,
        Patient Age At Diagnosis,
        Histological Classification Of Tumor,
        Surgical Procedure Type,
        Length Of Follow Up Years,
        Histological Diagnosis Skin Tumor,
        Disease Free Survival Months,
        Type Of Surgical Resection Primary Tumor,
        Surgical Procedures Performed,
        Metastatic Sites Reported,
        Granulocytic Sarcoma Anatomical Sites,
        Overall Survival After Surgery Months,
        Maximum Tumor Dimension Cm,
        Latency Years To Secondary Tumor,
        Imaging Modality For Diagnosis,
        Histologic Grade Of Tumor,
        Overall Survival Months From Diagnosis,
        Primary Hematologic Diagnosis,
        Primary Tumor Location 2,
        Tumor Anatomical Sites Involved,
        Disease Stage At Initial Diagnosis,
        Surgical Margin Status,
        Chemotherapy Agents Used,
        Kaposi Sarcoma Location,
        Patient Race Or Ethnicity,
        Years To Local Recurrence After Surgery,
        Therapeutic Agent Administered,
        Chemotherapy Cycles Administered,
        Preop Chemotherapy Agents,
        Hiv Serostatus,
        Total Radiation Dose Administered,
        Latency Period Radiation To Sarcoma,
        Cytogenetic Abnormality Identified,
        Histological Subtype Of Tumor,
        Figo Stage,
        Resected Tumor Weight Grams,
        Patient Cd4 T Cell Count,
        Tumor Dimensions Cm
    },
    yticklabel style={
        font=\sffamily\scriptsize,
        anchor=east,
        xshift=-2pt
    },
    xlabel={Abstracts with a reported extracted value (\%)},
    xlabel style={font=\sffamily\scriptsize},
    xmajorgrids=true,
    ymajorgrids=false,
    grid style={draw=gray!12, line width=0.3pt},
    axis line style={draw=gray!45, line width=0.4pt},
    tick style={draw=none},
    axis background/.style={fill=white},
    clip=true
]

\addplot[
    fill=coverageDark,
    draw=none,
    forget plot
] table[
    x expr={100*\thisrow{filled}/24577},
    y=id
] {\coveragedata};
\end{axis}
\end{tikzpicture}

\caption{Coverage of the retained CDEs in 24,577 unique
abstracts.}
\label{fig:cde-coverage}
\end{figure}

The distribution of CDEs per abstract varied widely across the 24,577 deduplicated abstracts (\autoref{fig:cde-coverage}). The most frequently populated CDEs were ``Patient sex'' (19,806 abstracts, 80.6\%), ``Patient age'' (17,917, 72.9\%), ``Age at diagnosis'' (17,770, 72.3\%), and ``Surgical resection performed'' (14,933, 60.8\%).

For ``Histological type of tumor'' (populated in 13,117 abstracts), the most frequent exact values were ``Leiomyosarcoma'' (2,038), ``Spindle cell sarcoma'' (1,392), ``Ewing sarcoma'' (1,197), and ``Osteogenic sarcoma'' (1,065). For the separate ``Primary tumor location'' CDE (populated in 10,926 abstracts), frequent labels included ``Uterus'' (850), ``Breast'' (734), ``Heart'' (699), and ``Lung'' (665).

A value for ``Treatment modalities administered'' was extracted in 8,402 abstracts, including the exact labels ``Surgery'' in 6,211 (73.9\%), ``Chemotherapy'' in 2,523 (30.0\%), and ``Radiotherapy'' in 1,663 (19.8\%). More specific labels, such as ``Surgical Excision'', remained separate. Among abstracts with an answer in the respective field, distant metastases were reported in 5,295 of 11,050 (47.9\%), tumour recurrence in 3,956 of 7,696 (51.4\%), and surgical resection in 13,397 of 14,933 (89.7\%). The outcome label ``Dead'' was recorded in 3,478 of 12,266 abstracts (28.4\%) with an outcome value.

Among answers meeting the parsing rules as described in the Methods, ``Patient age'' had a median of 46 years (IQR 26--63; $n=17{,}365$), and the separate ``Age at diagnosis'' a median of 47 years (IQR 26--63; $n=17{,}178$). The median maximum tumour dimension was 7~cm (IQR 4--12; $n=1{,}224$), the median number of chemotherapy cycles 5 (IQR 3--6; $n=504$), and the median radiation dose, among answers expressed in Gy, 50.4~Gy (IQR 45--60; $n=302$). \autoref{tab:numeric-summary} shows the field-specific denominators and excluded counts.

\begin{table}[htbp]
\centering
\caption{Descriptive summaries of five separate numeric CDEs in 24,577 unique abstracts.}
\label{tab:numeric-summary}
\small
\begin{tabularx}{\textwidth}{@{}>{\raggedright\arraybackslash}Xrrrr@{}}
\toprule
CDE (unit) & Non-empty & Included & Excluded & Median (IQR) \\
\midrule
Patient age (years) & 17,917 & 17,365 & 552 & 46 (26--63) \\
Age at diagnosis (years) & 17,770 & 17,178 & 592 & 47 (26--63) \\
Maximum tumour dimension (cm) & 2,578 & 1,224 & 1,354 & 7 (4--12) \\
Chemotherapy cycles & 757 & 504 & 253 & 5 (3--6) \\
Total radiation dose (Gy) & 448 & 302 & 146 & 50.4 (45--60) \\
\bottomrule
\end{tabularx}
\par\smallskip{\footnotesize\raggedright Included and excluded counts refer to non-empty answers passing or failing the field-specific parsing rules, respectively. Each row is a separate CDE, not a common complete-case cohort. Values were not combined across fields or converted between units. Patient age can refer to presentation, diagnosis, admission, or treatment, whereas age at diagnosis is a distinct question. These are summaries of uncorrected exported answers; numeric parseability does not establish correct units, tumour attribution, or treatment context. IQR: interquartile range.\par}
\end{table}

\section{Discussion}
\label{sec:discussion}

This study demonstrates an automated workflow in which an agentic system proposed a CDE schema before populating it through information extraction. We applied this workflow to titles and abstracts of PubMed-indexed sarcoma case reports retrieved using a predefined search strategy. The configured pipeline performed screening, schema induction, and information extraction automatically; investigators designed the workflow and evaluated its outputs.
While the approach could successfully be applied to a large corpus of sarcoma case reports of 24,588 abstracts, the expert review identified limitations in the generated schema with only 19 of 53 elements receiving approval from both reviewers on all three dimensions correctness, relevance, and clarity. 

The methodological contribution of this work lies in connecting question generation, semantic consolidation, response-vocabulary construction, and evidence-linked IE. Existing studies have so far shown that LLMs can extract predefined trial variables or organise case report information into predefined clinical categories, or populate a user-supplied schema \cite{caufieldStructuredPromptInterrogation2024}. Related work has used LLMs to map local variables onto existing NIH CDEs \cite{wangCDEMapperEnhancingNational2025} or to learn ontological structure \cite{babaeigiglouLLMs4OLLargeLanguage2023}. In our work, the information structure itself was derived from the literature under minimal investigator-defined constraints. The two-phase design of the workflow (CDE schema definition + subsequent information extraction) makes both the schema and its populated values available for inspection, allowing problems in either stage to be examined separately. 

The combination of agentic and direct inference stages within our workflow is also important: an agentic approach allowed the model to select actions and tool calls to propose and consolidate CDEs, whereas abstract screening and final information extraction used direct LLM calls. This follows the distinction between agents, which direct their own process, and workflows, which coordinate LLM calls through predefined code paths, as described by Anthropic (\url{https://www.anthropic.com/research/building-effective-agents}), and the argument that agent design should account for cost as well as accuracy \cite{kapoorAIAgentsThat2025}.

With regard to the quality of the CDE schema, we showed that correctly reproducing an answer within a given schema does not establish that the schema asks the right question or offers adequate response options. For example, the categorical tumour-dimensions CDE combines specific measurements with the descriptive label ``Large'' as allowed values. High agreement on the extracted value can therefore coexist with a lower approval of the schema. Reviewers assessed the suitability of the CDEs and their response options, whereas the answer-level evaluation measured agreement with reference values for those CDEs. Neither evaluation, however, establishes that the schema captures all clinically relevant concepts.

Interestingly, only 53 of the 1,570 master CDEs (3.4\%) were supported by at least 50 distinct source abstracts and retained for IE. This threshold selected commonly proposed concepts, but frequency alone does not establish clinical importance. Iseli et al. described ``insular criteria'' as important, relevant decision criteria used by only a minority of clinicians \cite{iseli2020insular}. Although our threshold measures occurrence in abstracts rather than use in clinical decisions, the concept highlights a possible limitation: uncommon features may matter greatly for a particular sarcoma subtype or unusual clinical situation. 

The 53 selected CDEs covered demographics, histology and pathology, primary and metastatic anatomy, stage and grade, surgery, systemic treatment, radiotherapy, recurrence, survival, and follow-up (Supplementary Table S2). However, despite two consolidation passes, the set retained several partially overlapping CDEs: three CDEs related to patient age (two numeric and one categorical with age-group labels such as ``40s'' or ``Neonate''), multiple CDEs regarding histology, and several CDEs regarding tumour location, two of which asked for the primary tumour location in different wording.

Evidence spans offer an additional but limited safeguard. They make extracted values traceable to the source text for non-empty answers. However, a matching span does not guarantee a correct value as it may refer to background information, a different tumour, a primary rather than a metastatic site, or a duration measured from the wrong starting point. Similar evidence-verification steps have been proposed for clinical extraction \cite{geroSelfverificationImprovesFewshot2023}, and LLM-cited sources often fail to support the claims they are attached to \cite{wuAutomatedFrameworkAssessing2025}.

A subsequent version of the pipeline should include a defined quality-assurance stage after schema generation and before routine reuse. Automated checks could identify duplicate questions, inconsistent response options or missing units. Clinical reviewers could then assess the questions and vocabularies most likely to affect interpretation.

The CDE schema also needs more precise representations of quantities and clinical context. Temporal relations remain a known difficulty in clinical text \cite{sunEvaluatingTemporalRelations2013}. Extracted findings should identify whether they concern the primary tumour, a recurrence, a metastasis, or background information. Future work should standardise the representation and downstream handling of explicitly negative findings, non-mention, and extraction failures across all CDEs, building on established work on negation detection in clinical NLP \cite{chapmanSimpleAlgorithmIdentifying2001}. These proposed improvements were not applied to the output analysed in this study.


The extracted data described a heterogeneous collection of case reports spanning diverse diagnoses, anatomical sites, treatments, and reported outcomes. Demographic information was frequently available, whereas quantitative details such as tumour dimensions and radiation doses were less consistently captured. Despite this uneven coverage, the structured output enabled descriptive summaries of a large literature corpus while preserving the individual data elements generated by the pipeline.

Automated extraction at this scale could support bibliometric analyses and hypothesis generation in rare cancers such as sarcoma \cite{stillerDescriptiveEpidemiologySarcomas2013}, where clinically unusual situations may be documented primarily in individual case reports, which are valued for detecting novel presentations and generating hypotheses but cannot support generalisation \cite{nissenClinicalCaseReport2014}. Combined with publication metadata, structured clinical information could help characterise which disease presentations and treatment approaches receive attention in the literature, examine changes over time, and identify underrepresented topics. It could also facilitate retrieval of uncommon clinical constellations for focused review and formulation of research hypotheses.

Several further limitations should be noted. First, our approach relied on abstracts rather than full case reports. Abstracts are concise by design and may omit clinically relevant details, such as treatment specifics, follow-up duration, or histopathological findings \cite{cohenStructuralContentAspects2010}. Second, publication records were not linked to verified unique patients, so the same patient may appear in more than one report and be overrepresented in the resulting dataset. Duplicate and overlapping publication of the same patients is a recognised problem in evidence synthesis \cite{vonelmDifferentPatternsDuplicate2004}. Last, annotators assessed only model outputs, which can introduce anchoring and automation bias \cite{goddardAutomationBiasSystematic2012}.
Because empty pipeline responses were not reviewed for missed information, the evaluation does not establish extraction completeness or recall. Some field-specific agreement estimates were based on very few comparisons, limiting the conclusions that can be drawn about extraction performance for those CDEs. The evaluation sample was drawn from the same corpus used for schema induction and therefore does not, by itself, establish generalisation to an independent corpus.

Applying the pipeline to full-text reports, where licensing permits, would let us capture a richer and more complete set of CDEs. Record linkage based on patient characteristics, authorship, and institutional affiliation could help identify and merge duplicate cases and could also be implemented using agentic approaches. Future work should also include a human-in-the-loop step for schema validation. Domain experts could review and refine the proposed CDE schema before large-scale extraction. That review would catch inconsistencies such as the type conflicts described above early, and it would make the standardised output more clinically valid.


In this study, we applied LLM agents to derive a CDE schema from 24,588 sarcoma case report abstracts, and constrained LLM calls to extract values for the 53 most frequently proposed CDEs across the corpus. Answer-level agreement was high within the evaluation sample, but only 19 of 53 retained CDEs received approval from both reviewers on all three schema-quality criteria. These findings identify schema quality as a major limitation of the current workflow. Agent-generated schemas require expert curation and automated checks for overlapping fields, inadequate value sets, and inconsistent representations before large-scale extraction.

\section{Methods}
\label{sec:methods}
\subsection{Study design}

To investigate whether LLM agents can derive a useful CDE schema from biomedical literature, we developed a constrained agentic workflow, in which the model plans its own steps and calls tools \cite{wangSurveyLargeLanguage2024} and applied it to sarcoma case report abstracts. We retrieved and screened PubMed records, proposed CDEs independently from each eligible abstract, and consolidated them into a corpus-level schema. We then used direct LLM calls to extract values for the most frequently proposed CDEs across the corpus. Domain experts evaluated the retained schema and a sample of extracted answers. Finally, we characterised the information represented in the extracted corpus.

\subsection{Corpus creation}

We conducted a retrospective study using titles and abstracts of sarcoma case reports indexed in PubMed. First, we constructed the input corpus through classification of abstracts and used an agentic workflow to derive a CDE schema. Next, we attempted to extract values for CDEs supported by at least 50 distinct source abstracts from every abstract in the input corpus, requiring supporting text spans for non-empty answers. Finally, human domain experts evaluated the retained CDE schema and a sample of extracted answers.

PubMed was searched on 21 May 2026 with the following query, which combines sarcoma MeSH terms and title/abstract terms with the publication type ``Case Reports'' and restricts publication dates to 30 April 2026 or earlier:
\begin{quote}\small\ttfamily
(("Sarcoma"[Mesh] OR "Sarcoma, Soft Tissue"[Mesh] OR "Epithelioid Sarcoma"[Mesh] OR "Sarcoma, Spindle Cell"[Mesh]) OR (sarcoma*[Title/Abstract] OR "soft tissue sarcoma*"[Title/Abstract] OR "epithelioid sarcoma*"[Title/Abstract] OR "spindle cell sarcoma*"[Title/Abstract])) AND "Case Reports"[Publication Type] AND ("1000/01/01"[Date - Publication] : "2026/04/30"[Date - Publication])
\end{quote}
Title, abstract, journal, and publication year were retrieved for each record, and records without an abstract were excluded. An LLM-based classifier then assessed each remaining abstract. It received the PMID, title, and abstract and returned a structured decision (case report or not, exclusion reason, and rationale). It was instructed to include only single-patient human case reports and to exclude case series of two or more patients, cohorts, trials, reviews, guidelines, methods papers, non-human studies, and abstracts that did not clearly describe a single patient.

\subsection{CDE schema induction}

For each eligible abstract, an extraction agent proposed CDEs. Each abstract was processed in a separate conversation, so the agent kept no memory of previous abstracts. The agent retrieved the abstract, recorded each proposed CDE with a separate function call, and then closed the abstract. Each proposed CDE consisted of a question, a type, a description, and the value found in the abstract. The question had to be a closed question answerable with yes/no, a number, or one discrete category. Two CDE types from the NIH definitions were permitted: categorical elements (NIH ``Value List'') for qualitative attributes with discrete answer options, and numeric elements (NIH ``Number'') for quantities, with the unit stated in the description. The runtime rejected CDEs whose question did not end with a question mark or whose value was empty, and returned the error to the agent for correction. 

Next, CDEs with identical question text and type were merged. Each merged CDE took the most frequent description among its members, categorical CDEs received the union of their members' values as their value set, and the identifiers of all member CDEs were kept as provenance.

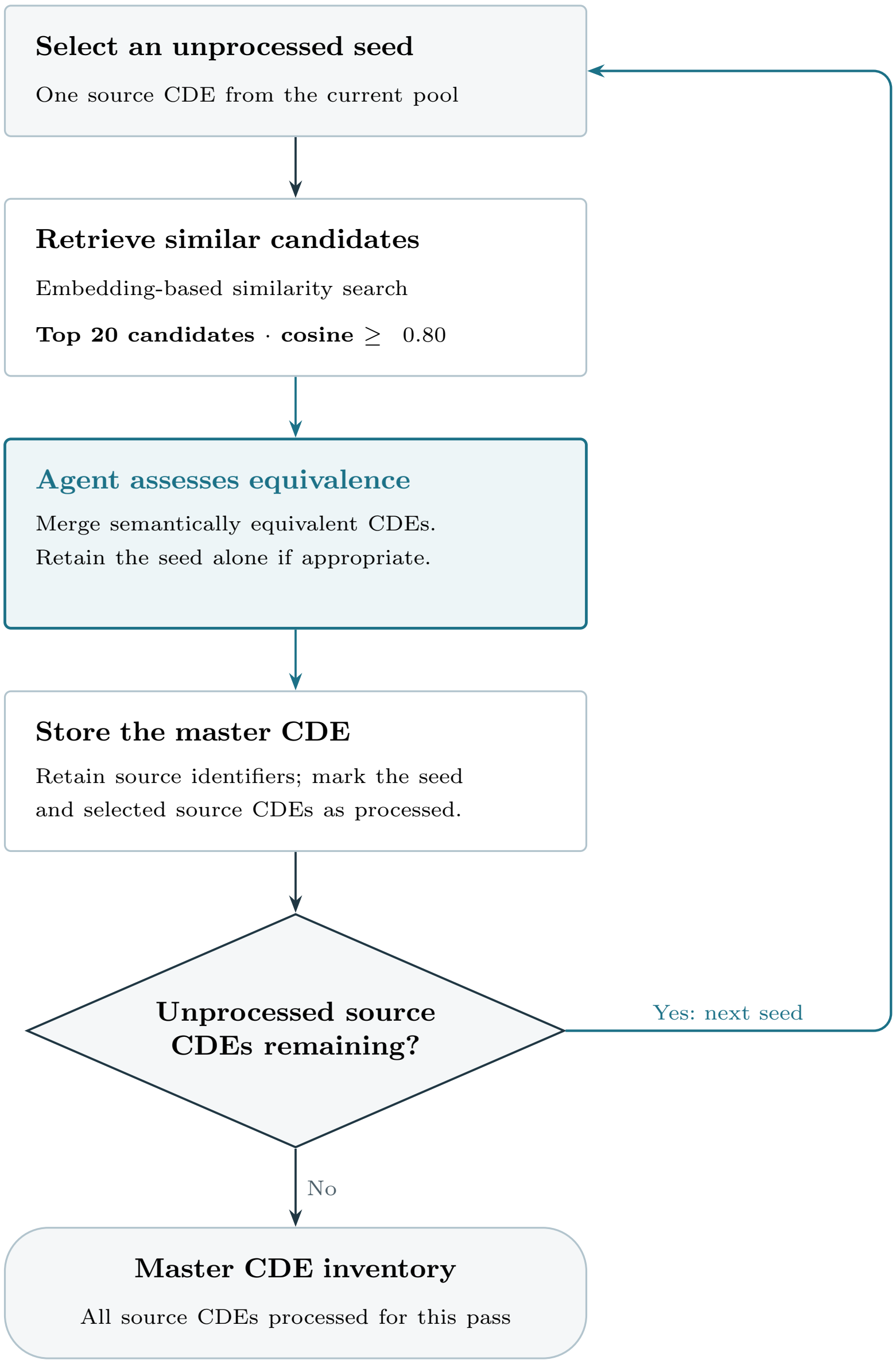
\begin{figure}[htbp]
\centering
\begin{tikzpicture}[
    font=\small,
    node distance=7mm,
    >={Stealth[length=2mm]},
    box/.style={
        rectangle,
        rounded corners=2pt,
        draw=border,
        fill=light,
        line width=0.6pt,
        text width=6cm,
        align=left,
        inner sep=3.5mm
    },
    agent/.style={
        box,
        draw=accent,
        fill=agentfill,
        line width=0.9pt
    },
    decision/.style={
        diamond,
        aspect=2.3,
        draw=ink,
        fill=light,
        line width=0.7pt,
        text width=4cm,
        align=center,
        inner sep=1mm,
        font=\small\bfseries
    },
    terminal/.style={
        box,
        rounded corners=5mm,
        align=center
    },
    flow/.style={
        ->,
        draw=ink,
        line width=0.7pt
    },
    loop/.style={
        ->,
        draw=accent,
        line width=0.8pt,
        rounded corners=2mm
    }
]

\definecolor{ink}{HTML}{203743}
\definecolor{border}{HTML}{B2C4CD}
\definecolor{light}{HTML}{F5F7F8}
\definecolor{accent}{HTML}{1C7187}
\definecolor{agentfill}{HTML}{EDF5F7}

\node[box] (seed) {
    {\bfseries Select an unprocessed seed}\\[1.5mm]
    {\footnotesize One source CDE from the current pool}
};

\node[box, below=of seed, fill=white] (retrieve) {
    {\bfseries Retrieve similar candidates}\\[1.5mm]
    {\footnotesize Embedding-based similarity search}\\[1.5mm]
    {\footnotesize\bfseries
        Top 20 candidates $\cdot$ cosine $\geq 0.80$}
};

\node[agent, below=of retrieve] (assess) {
    {\bfseries\color{accent} Agent assesses equivalence}\\[1mm]
    {\footnotesize
        Merge semantically equivalent CDEs.\\
        Retain the seed alone if appropriate.\\
       }
};

\node[box, below=of assess, fill=white] (store) {
    {\bfseries Store the master CDE}\\[1mm]
    {\footnotesize
        Retain source identifiers; mark the seed\\
        and selected source CDEs as processed.}
};

\node[decision, below=of store] (remaining) {
    Unprocessed source\\
    CDEs remaining?
};

\node[terminal, below=9mm of remaining] (inventory) {
    {\bfseries Master CDE inventory}\\[1.5mm]
    {\footnotesize All source CDEs processed for this pass}
};

\draw[flow] (seed) -- (retrieve);
\draw[flow, draw=accent] (retrieve) -- (assess);
\draw[flow, draw=accent] (assess) -- (store);
\draw[flow] (store) -- (remaining);

\draw[flow] (remaining.south)
    -- node[right, font=\footnotesize, text=ink!80] {No}
    (inventory.north);

\coordinate (loopright) at ($(seed.east)+(3.5cm,0)$);

\draw[loop]
    (remaining.east)
    -- node[above, font=\footnotesize, text=accent]
        {Yes: next seed}
    (remaining.east -| loopright)
    -- (loopright)
    -- (seed.east);

\end{tikzpicture}
\caption{Seed-based consolidation within each pass. For an unprocessed source CDE, embedding-based retrieval provided up to 20 candidates meeting a cosine-similarity threshold of 0.80 as input to the agent. The agent assessed semantic equivalence and created a master CDE from the seed and selected candidates, or retained the seed alone. Source identifiers were preserved, and the contributing source CDEs were marked as processed. The cycle continued until all source CDEs were processed for that pass; this stopping rule did not imply that no further consolidation was possible. The recorded first and second passes yielded 2,183 and 1,570 master CDEs, respectively.}
\label{fig:consolidation_detail}
\end{figure}

In two consolidation passes, an agent merged CDEs that described the same concept in different wording (Figure~\ref{fig:consolidation_detail}). Each CDE was embedded based on its question and description text. For each unprocessed CDE, the agent ran a similarity search that returned up to 20 unassigned candidates of the same type with a cosine similarity of at least 0.80. The agent then decided which candidates to merge into a new master CDE. It was instructed to favour precision over recall and to merge only candidates that asked for the same clinical variable and, for categorical CDEs, whose values were valid responses to the CDE’s question. Each input CDE ended up in exactly one master CDE. The agent composed the value set of a categorical CDE itself from the values of the merged candidate CDEs. A second consolidation pass used the master CDEs from the first pass as input. Each resulting master CDE could be traced to all its source abstracts.

Next, another agent standardised the value set of each categorical master CDE separately. It returned a revised list of labels and could record relationships between labels, either as synonyms or as subtypes. The instructions were:
\begin{itemize}
\item split a label only if it clearly encoded several independent answers;
\item keep named regimens, anatomical sites, dimensions, and single combined findings intact;
\item normalise lightly, using ``Yes'' and ``No'' for yes/no questions;
\item remove case-insensitive duplicates;
\item keep every distinct clinical answer.
\end{itemize}
The runtime rejected any value set containing a label with a comma, semicolon, slash, vertical bar, ``+'', ``and'', or ``or''. 
If no valid value set was saved, the original labels were kept. 
Relationships between labels were stored in the schema but not used to collapse labels during IE or analysis.

The support of a master CDE was defined as the number of distinct source abstracts among its proposed CDEs. Master CDEs with a support of at least 50 abstracts were retained for IE and evaluation. This criterion counts how often the concept was proposed during schema creation, not the number of non-empty answers after IE. The threshold was chosen for computational feasibility: IE requires one model request per abstract--CDE pair, so retaining the 154 master CDEs supported by at least 10 abstracts would have required approximately 3.8 million requests, compared with approximately 1.3 million for the 53 CDEs retained at the 50-abstract threshold.

\subsection{Data extraction}

The retained 53 CDEs were filled for each accepted abstract with direct LLM calls, one request per abstract--CDE pair. An initial agent-based implementation was abandoned because of its processing overhead. Each request contained the PMID, title, abstract, the CDE's identifier, question, type, and description, and, for categorical CDEs, the complete value set. Numeric answers were returned as the reported value or as “unknown”. Categorical answers consisted of zero or more labels from the value set, never joined into one string.

The runtime checked each answer:
\begin{itemize}
\item numeric answers had to be non-empty, but their content was not validated as numeric;
\item each selected label had to pass the same atomicity check as in standardisation and match a value-set entry, ignoring case;
\item non-empty answers required an evidence span that occurred in the abstract, ignoring case and whitespace.
\end{itemize}
A rejected answer was returned to the LLM together with the validation error, for up to three attempts per pair. CDEs that never passed validation were left out of the abstract's record and logged with their last validation error. Failed extraction can therefore be distinguished from non-mention, which was stored as ``unknown'' for numeric CDEs or as an empty value list for categorical CDEs.

\subsection{Implementation details}

All generative steps (screening, CDE proposal, consolidation, standardisation, and IE) used gpt-oss-120b (OpenAI) \cite{openai2025gptoss}, an open-weight LLM that performed well in our previous study of radiotherapy case reports \cite{dennstaedt2025classification}, with default sampling parameters. Embeddings were generated with embeddinggemma-300m (768 dimensions). AG2 [https://github.com/ag2ai/ag2] served as the agent framework. The models ran on two Mac Studio systems (512~GB unified memory each), with requests distributed through LiteLLM behind an OpenAI-compatible API. All model outputs used JSON schemas with strict validation. Transient errors, such as rate limits, server errors, and connection failures, were retried up to four times with exponential backoff, separately from the validation retries described above. Every stage wrote checkpoints regularly, could resume from its last output, and stored provenance and failures in structured files. Recorded token usage was 21.3 million tokens for screening, 259.0 million for CDE proposal, and 1,159.4 million for IE; consolidation and standardisation did not record token counts.

\subsection{Expert evaluation of the CDE schema}
Two medical professionals and researchers (MN and AB) independently evaluated the retained CDEs.
For each CDE, reviewers independently assessed correctness, relevance, and clarity. Correctness concerned the clinical and conceptual validity of the CDE, including consistency between its question, description, data type, and, where applicable, permissible values (= value set). Relevance concerned whether the CDE captured information useful for describing a sarcoma case report. Clarity concerned whether the question and response options were understandable and unambiguous. 

For schema evaluation, we summarised reviewer-specific approval and approval by both reviewers separately for correctness, relevance, and clarity. Inter-reviewer agreement included both concordant approval and concordant non-approval and was quantified using percentage agreement and Cohen's $\kappa$. 

\subsection{Expert evaluation of extraction performance}
For a random sample of 100 abstracts (sampled from the entire corpus of 24,588 case reports), the same two reviewers independently assessed whether each non-empty extracted answer correctly answered its CDE question based on the source abstract. If an answer was considered incorrect, reviewers provided a corrected answer. Disagreements concerning the correct answer were adjudicated by a third medical professional (KA). Empty pipeline responses were not reviewed for missed information; the evaluation therefore assessed the correctness of extracted answers rather than the completeness of extraction. Exact agreement was calculated as the number of matching answers divided by the number of abstract--CDE pairs with a non-empty answer in the pipeline output, the reference, or both. Jointly empty pairs were excluded. The 95\% confidence interval of the exact agreement was estimated by bootstrap resampling at the abstract level to account for the grouping of CDE evaluations within abstracts. Non-matches were classified as pipeline-only answers, reference-only answers, or different non-empty answers.
The evaluation methodology was based on the methods used in previous studies of LLM-based IE from biomedical literature \cite{dennstaedt2026trialclassification,windisch2024metastatic,dennstaedt2025classification}.

\subsection{Descriptive analysis of extracted information}
The descriptive analysis was retrospective and exploratory. The unit of analysis was the abstract, not an independently verified patient; proportions therefore describe the content of abstracts reporting a value and are not estimates of incidence, prevalence, treatment use, or prognosis.

Continuous variables (i.e., coverage across fields and the number of filled fields per abstract) were described by median and interquartile ranges (IQRs). Coverage was defined as the proportion of abstracts with a non-empty answer. Explicit labels such as ``Not specified'' were kept as pipeline responses. Empty answers can reflect either non-mention in the abstract or a failed extraction and were not coded as negative findings. Coverage therefore measures non-empty pipeline output rather than verified availability of the corresponding clinical information.
Furthermore, we selected five numeric CDEs as illustrative examples of the quantitative information produced by the pipeline: two distinct age fields, maximum tumour dimension, chemotherapy cycles, and total radiation dose. Together, they cover demographic, tumour, and treatment information while allowing us to examine how answer format affects numeric analysis. They were analysed separately and were not intended as a comprehensive summary of all 14 numeric CDEs. Only answers consisting of a single non-negative number with an optional recognised unit were included. The accepted units were years for age, centimetres for tumour dimension, and cycles for chemotherapy; bare numbers were read in the unit stated by the question. Radiation dose required an explicit ``Gy''. No unit conversion or imputation was applied. Categorical fields were described as counts and percentages of exact labels with field-specific denominators.

\FloatBarrier


\section*{Data availability}
The datasets generated and/or analysed during the current study are available in the Zenodo repository, 10.5281/zenodo.23371058.

\section*{Code availability}
The source code for this publication is available in the Zenodo repository, 10.5281/zenodo.23371058.


\bibliography{references}


\backmatter

\bmhead{Acknowledgements}
We thank Tomy Jefin and Pascal Ros Chassignet, who first explored possibilities for automated information extraction from sarcoma case reports as part of their research internship at the University Hospital of Bern in the research group of Fabio Dennstädt. This study received no funding.

\bmhead{Author contributions}

DR and FD conceived and designed the study. DR developed the software pipeline, performed the data retrieval, screening, schema creation, and data extraction. 
FD contributed clinical expertise during the pipeline development process.
IF and FD performed the statistical analysis. 
MN and AB independently evaluated the 53 retained CDEs and the extracted values in the validation sample.
KA adjudicated disagreements in the extraction evaluation. 
DR provided the computing infrastructure. 
DR and FD drafted the manuscript. 
All authors critically revised the manuscript for important intellectual content, approved the final version, and agree to be accountable for all aspects of the work.

\bmhead{Competing interests}
KA is employed by F. Hoffmann-La Roche AG. 
All other authors declare no financial or non-financial competing interests.



\end{document}


\thispagestyle{plain}
\begin{center}
{\Large\bfseries Supplementary Information}\\[8pt]
{\large Agentic induction of common data elements and information extraction from sarcoma case report abstracts}\\[6pt]
\end{center}
\vspace{4pt}

\noindent This document contains three supplementary tables. Numbered IDs 1--53 refer to the same retained common data elements (CDEs) in all tables; IDs 1--14 are numeric and IDs 15--53 categorical. Field names are the CDE identifiers in title case, and questions are reproduced unchanged from the schema. Coverage and label counts refer to the 24,577 unique abstracts used for the descriptive analyses.

\begin{itemize}\setlength{\itemsep}{2pt}
\item \textbf{Supplementary Table~\ref{supp:evaluation}} reports the answer-level evaluation of the 100-abstract validation sample for every retained CDE.
\item \textbf{Supplementary Table~\ref{supp:schema}} lists the retained schema: each CDE's question, type, support in the source abstracts, number of permitted labels, and coverage.
\item \textbf{Supplementary Table~\ref{supp:labels}} lists up to five of the most frequent exact labels for each categorical CDE.
\end{itemize}

\listoftables

\clearpage
\begingroup
\footnotesize
\setlength{\tabcolsep}{3pt}
\begin{longtable}{r >{\raggedright\arraybackslash}p{7.5cm} r r r r r r}
\caption{Answer-level evaluation for all 53 retained CDEs.}\label{supp:evaluation}\\
\toprule
ID & Field & Match & MO & RO & Diff. & $n$ & \% \\
\midrule\endfirsthead
\toprule
ID & Field & Match & MO & RO & Diff. & $n$ & \% \\
\midrule\endhead
\midrule\multicolumn{8}{r}{Continued on next page}\\\endfoot
\bottomrule\endlastfoot
1 & Chemotherapy Cycles Administered & 4 & 0 & 0 & 0 & 4 & 100.0 \\
2 & Years To Local Recurrence After Surgery & 4 & 0 & 0 & 1 & 5 & 80.0 \\
3 & Latency Years To Secondary Tumor & 7 & 1 & 0 & 2 & 10 & 70.0 \\
4 & Disease Free Survival Months & 18 & 0 & 0 & 3 & 21 & 85.7 \\
5 & Length Of Follow Up Years & 20 & 0 & 0 & 0 & 20 & 100.0 \\
6 & Maximum Tumor Dimension Cm & 8 & 0 & 0 & 0 & 8 & 100.0 \\
7 & Age At Diagnosis & 73 & 0 & 0 & 1 & 74 & 98.6 \\
8 & Patient Age Years & 74 & 0 & 0 & 1 & 75 & 98.7 \\
9 & Patient Cd4 T Cell Count & 1 & 0 & 0 & 0 & 1 & 100.0 \\
10 & Resected Tumor Weight Grams & 0 & 0 & 0 & 0 & 0 & -- \\
11 & Latency Period Radiation To Sarcoma & 3 & 0 & 0 & 0 & 3 & 100.0 \\
12 & Overall Survival After Surgery Months & 10 & 0 & 0 & 1 & 11 & 90.9 \\
13 & Overall Survival Months From Diagnosis & 8 & 0 & 0 & 2 & 10 & 80.0 \\
14 & Total Radiation Dose Administered & 1 & 0 & 0 & 0 & 1 & 100.0 \\
15 & Did Patient Develop Distant Metastases & 44 & 0 & 0 & 2 & 46 & 95.7 \\
16 & Tumor Recurrence During Follow Up & 37 & 0 & 0 & 0 & 37 & 100.0 \\
17 & Was Surgical Resection Performed & 54 & 1 & 0 & 2 & 57 & 94.7 \\
18 & Granulocytic Sarcoma Anatomical Sites & 11 & 5 & 0 & 0 & 16 & 68.8 \\
19 & Tumor Dimensions Cm & 0 & 0 & 0 & 0 & 0 & -- \\
20 & Cytogenetic Abnormality Identified & 4 & 0 & 0 & 0 & 4 & 100.0 \\
21 & Imaging Modality For Diagnosis & 7 & 0 & 0 & 0 & 7 & 100.0 \\
22 & Figo Stage & 2 & 0 & 0 & 0 & 2 & 100.0 \\
23 & Histologic Grade Of Tumor & 13 & 0 & 0 & 0 & 13 & 100.0 \\
24 & Histological Classification Of Tumor & 25 & 0 & 0 & 0 & 25 & 100.0 \\
25 & Histological Diagnosis Skin Tumor & 19 & 0 & 0 & 0 & 19 & 100.0 \\
26 & Tumor Histological Pattern & 54 & 1 & 0 & 0 & 55 & 98.2 \\
27 & Histological Subtype Of Tumor & 2 & 0 & 0 & 0 & 2 & 100.0 \\
28 & Histological Type Of Tumor & 48 & 3 & 0 & 0 & 51 & 94.1 \\
29 & Histopathologic Diagnosis Biopsy & 50 & 0 & 0 & 0 & 50 & 100.0 \\
30 & Patient Age At Diagnosis & 17 & 0 & 0 & 1 & 18 & 94.4 \\
31 & Patient Race Or Ethnicity & 10 & 0 & 0 & 0 & 10 & 100.0 \\
32 & Patient Sex & 75 & 2 & 0 & 0 & 77 & 97.4 \\
33 & Hiv Serostatus & 2 & 0 & 0 & 0 & 2 & 100.0 \\
34 & Primary Hematologic Diagnosis & 9 & 0 & 0 & 0 & 9 & 100.0 \\
35 & Metastatic Sites Reported & 10 & 0 & 0 & 0 & 10 & 100.0 \\
36 & Surgical Procedures Performed & 15 & 1 & 0 & 0 & 16 & 93.8 \\
37 & Therapeutic Agent Administered & 4 & 0 & 0 & 0 & 4 & 100.0 \\
38 & Treatment Modalities Administered & 34 & 0 & 0 & 0 & 34 & 100.0 \\
39 & Primary Treatment Type & 38 & 0 & 0 & 0 & 38 & 100.0 \\
40 & Surgical Procedure Type & 21 & 0 & 0 & 0 & 21 & 100.0 \\
41 & Type Of Surgical Resection Primary Tumor & 14 & 0 & 0 & 0 & 14 & 100.0 \\
42 & Primary Tumor Location 2 & 4 & 0 & 0 & 0 & 4 & 100.0 \\
43 & Disease Stage At Initial Diagnosis & 6 & 0 & 0 & 0 & 6 & 100.0 \\
44 & Final Clinical Outcome & 48 & 1 & 0 & 0 & 49 & 98.0 \\
45 & Surgical Margin Status & 7 & 0 & 0 & 0 & 7 & 100.0 \\
46 & Extramedullary Tumor Location & 42 & 4 & 0 & 0 & 46 & 91.3 \\
47 & Kaposi Sarcoma Location & 3 & 2 & 0 & 0 & 5 & 60.0 \\
48 & Metastatic Tumor Location & 18 & 4 & 0 & 2 & 24 & 75.0 \\
49 & Primary Lesion Location & 27 & 7 & 1 & 4 & 39 & 69.2 \\
50 & Primary Tumor Location & 45 & 2 & 0 & 0 & 47 & 95.7 \\
51 & Tumor Anatomical Sites Involved & 5 & 0 & 0 & 0 & 5 & 100.0 \\
52 & Chemotherapy Agents Used & 2 & 0 & 0 & 0 & 2 & 100.0 \\
53 & Preop Chemotherapy Agents & 2 & 0 & 0 & 0 & 2 & 100.0 \\
\end{longtable}
\noindent MO: model-only value; RO: reference-only value; Diff.: different non-empty answers. The denominator n includes matches and all three error categories. A dash denotes no non-empty comparisons. Numeric IDs are display indices; stable field identifiers are provided in the accompanying schema dictionary.
\par\smallskip
\noindent Field-specific percentages are based on the comparison counts shown in column $n$. High agreement based on very few comparisons provides limited evidence about extraction performance. Empty pipeline responses were not systematically assessed for missed information, so these results do not establish extraction completeness.
\endgroup

\clearpage
\begingroup
\small
\setlength{\tabcolsep}{3pt}
\begin{longtable}{r >{\raggedright\arraybackslash}p{7.0cm} >{\raggedright\arraybackslash}p{1.6cm} r r r}
\caption{Retained schema and field coverage in 24,577 unique abstracts.}\label{supp:schema}\\
\toprule
ID & Field and unchanged question & Type & Source $n$ & Labels & Filled $n$ (\%) \\
\midrule\endfirsthead
\toprule
ID & Field and unchanged question & Type & Source $n$ & Labels & Filled $n$ (\%) \\
\midrule\endhead
\midrule\multicolumn{6}{r}{Continued on next page}\\\endfoot
\bottomrule\endlastfoot
1 & \textbf{Chemotherapy Cycles Administered}\newline How many cycles of chemotherapy were administered? & Numeric & 51 & -- & 757 (3.1\%) \\
2 & \textbf{Years To Local Recurrence After Surgery}\newline How many years after initial surgery did the tumor recur locally? & Numeric & 51 & -- & 996 (4.1\%) \\
3 & \textbf{Latency Years To Secondary Tumor}\newline How many years after the primary cancer treatment or diagnosis did a secondary tumor develop? & Numeric & 83 & -- & 2,467 (10.0\%) \\
4 & \textbf{Disease Free Survival Months}\newline What is the duration of disease-free survival (months)? & Numeric & 139 & -- & 4,660 (19.0\%) \\
5 & \textbf{Length Of Follow Up Years}\newline What is the length of follow-up reported (in years)? & Numeric & 100 & -- & 4,801 (19.5\%) \\
6 & \textbf{Maximum Tumor Dimension Cm}\newline What is the maximum tumor dimension in centimeters? & Numeric & 2,072 & -- & 2,578 (10.5\%) \\
7 & \textbf{Age At Diagnosis}\newline What is the patient's age at diagnosis (years)? & Numeric & 11,231 & -- & 17,770 (72.3\%) \\
8 & \textbf{Patient Age Years}\newline What is the patient's age in years at presentation, diagnosis, admission, or treatment? & Numeric & 7,106 & -- & 17,917 (72.9\%) \\
9 & \textbf{Patient Cd4 T Cell Count}\newline What is the patient's CD4+ T-cell count? & Numeric & 51 & -- & 73 (0.3\%) \\
10 & \textbf{Resected Tumor Weight Grams}\newline What is the weight of the resected tumor specimen in grams? & Numeric & 53 & -- & 159 (0.6\%) \\
11 & \textbf{Latency Period Radiation To Sarcoma}\newline What was the latency period (in years) between radiation therapy (or breast conserving therapy) and diagnosis of a subsequent sarcoma? & Numeric & 66 & -- & 434 (1.8\%) \\
12 & \textbf{Overall Survival After Surgery Months}\newline What was the overall survival after surgery in months? & Numeric & 59 & -- & 2,774 (11.3\%) \\
13 & \textbf{Overall Survival Months From Diagnosis}\newline What was the overall survival time (in months) from cancer diagnosis to death? & Numeric & 183 & -- & 1,935 (7.9\%) \\
14 & \textbf{Total Radiation Dose Administered}\newline What was the total radiation dose administered? & Numeric & 162 & -- & 448 (1.8\%) \\
15 & \textbf{Did Patient Develop Distant Metastases}\newline Did the patient develop distant metastases? & Categorical & 113 & 2 & 11,050 (45.0\%) \\
16 & \textbf{Tumor Recurrence During Follow Up}\newline Did the patient experience tumor recurrence during postoperative follow-up? & Categorical & 92 & 2 & 7,696 (31.3\%) \\
17 & \textbf{Was Surgical Resection Performed}\newline Was surgical resection performed? & Categorical & 51 & 2 & 14,933 (60.8\%) \\
18 & \textbf{Granulocytic Sarcoma Anatomical Sites}\newline What anatomical sites are involved by granulocytic sarcoma lesions? & Categorical & 55 & 67 & 2,982 (12.1\%) \\
19 & \textbf{Tumor Dimensions Cm}\newline What are the dimensions of the primary tumor (length $\times$ width $\times$ height) in centimeters? & Categorical & 126 & 5 & 43 (0.2\%) \\
20 & \textbf{Cytogenetic Abnormality Identified}\newline What cytogenetic abnormality (e.g., translocation, gene fusion, amplification) was identified in the tumor? & Categorical & 206 & 82 & 419 (1.7\%) \\
21 & \textbf{Imaging Modality For Diagnosis}\newline What imaging modality was used to confirm the diagnosis of the tumor? & Categorical & 94 & 40 & 2,090 (8.5\%) \\
22 & \textbf{Figo Stage}\newline What is the FIGO stage of the tumor? & Categorical & 67 & 18 & 210 (0.9\%) \\
23 & \textbf{Histologic Grade Of Tumor}\newline What is the histologic grade of the tumor? & Categorical & 509 & 16 & 1,969 (8.0\%) \\
24 & \textbf{Histological Classification Of Tumor}\newline What is the histological classification of the tumor? & Categorical & 96 & 9 & 4,919 (20.0\%) \\
25 & \textbf{Histological Diagnosis Skin Tumor}\newline What is the histological diagnosis of the skin tumor? & Categorical & 53 & 22 & 4,717 (19.2\%) \\
26 & \textbf{Tumor Histological Pattern}\newline What is the histological pattern or diagnosis of the tumor? & Categorical & 130 & 116 & 11,985 (48.8\%) \\
27 & \textbf{Histological Subtype Of Tumor}\newline What is the histological subtype of the tumor? & Categorical & 4,602 & 2 & 308 (1.3\%) \\
28 & \textbf{Histological Type Of Tumor}\newline What is the histological type of the tumor? & Categorical & 106 & 40 & 13,117 (53.4\%) \\
29 & \textbf{Histopathologic Diagnosis Biopsy}\newline What is the histopathologic diagnosis based on the biopsy? & Categorical & 68 & 46 & 11,604 (47.2\%) \\
30 & \textbf{Patient Age At Diagnosis}\newline What is the patient's age at diagnosis? & Categorical & 232 & 72 & 5,132 (20.9\%) \\
31 & \textbf{Patient Race Or Ethnicity}\newline What is the patient's race or ethnicity? & Categorical & 419 & 45 & 1,057 (4.3\%) \\
32 & \textbf{Patient Sex}\newline What is the patient's sex? & Categorical & 17,429 & 4 & 19,806 (80.6\%) \\
33 & \textbf{Hiv Serostatus}\newline What is the patient's HIV serostatus? & Categorical & 308 & 9 & 675 (2.7\%) \\
34 & \textbf{Primary Hematologic Diagnosis}\newline What is the primary hematologic diagnosis? & Categorical & 76 & 29 & 1,755 (7.1\%) \\
35 & \textbf{Metastatic Sites Reported}\newline What metastatic sites are reported? & Categorical & 67 & 58 & 3,084 (12.5\%) \\
36 & \textbf{Surgical Procedures Performed}\newline What surgical procedures were performed? & Categorical & 92 & 46 & 3,349 (13.6\%) \\
37 & \textbf{Therapeutic Agent Administered}\newline What therapeutic agent was administered to the patient? & Categorical & 136 & 63 & 892 (3.6\%) \\
38 & \textbf{Treatment Modalities Administered}\newline What treatment modalities were administered to the patient? & Categorical & 303 & 30 & 8,402 (34.2\%) \\
39 & \textbf{Primary Treatment Type}\newline What type of primary treatment was administered? & Categorical & 59 & 11 & 9,306 (37.9\%) \\
40 & \textbf{Surgical Procedure Type}\newline What type of surgical procedure was performed for treatment? & Categorical & 1,007 & 28 & 4,877 (19.8\%) \\
41 & \textbf{Type Of Surgical Resection Primary Tumor}\newline What type of surgical resection was performed for the primary tumor? & Categorical & 119 & 100 & 3,471 (14.1\%) \\
42 & \textbf{Primary Tumor Location 2}\newline What was the anatomical location of the primary tumor? & Categorical & 809 & 3 & 1,680 (6.8\%) \\
43 & \textbf{Disease Stage At Initial Diagnosis}\newline What was the disease stage at initial diagnosis? & Categorical & 286 & 19 & 1,417 (5.8\%) \\
44 & \textbf{Final Clinical Outcome}\newline What was the final clinical outcome for the patient? & Categorical & 65 & 2 & 12,266 (49.9\%) \\
45 & \textbf{Surgical Margin Status}\newline What was the surgical margin status? & Categorical & 202 & 3 & 1,365 (5.6\%) \\
46 & \textbf{Extramedullary Tumor Location}\newline Where is the extramedullary tumor located? & Categorical & 70 & 67 & 9,884 (40.2\%) \\
47 & \textbf{Kaposi Sarcoma Location}\newline Where is the Kaposi sarcoma lesion located? & Categorical & 83 & 83 & 1,249 (5.1\%) \\
48 & \textbf{Metastatic Tumor Location}\newline Where is the metastatic tumor located? & Categorical & 622 & 140 & 5,397 (22.0\%) \\
49 & \textbf{Primary Lesion Location}\newline Where is the primary lesion located? & Categorical & 345 & 231 & 10,938 (44.5\%) \\
50 & \textbf{Primary Tumor Location}\newline Where is the primary tumor located? & Categorical & 17,296 & 79 & 10,926 (44.5\%) \\
51 & \textbf{Tumor Anatomical Sites Involved}\newline Which anatomical sites are involved by the tumor? & Categorical & 226 & 5 & 1,421 (5.8\%) \\
52 & \textbf{Chemotherapy Agents Used}\newline Which chemotherapy agent(s) were used for treatment? & Categorical & 364 & 140 & 1,271 (5.2\%) \\
53 & \textbf{Preop Chemotherapy Agents}\newline Which chemotherapy agents were used in the pre-operative (neoadjuvant) regimen? & Categorical & 52 & 47 & 721 (2.9\%) \\
\end{longtable}
\noindent Source $n$ is the number of distinct source PMIDs among the proposed CDEs assigned to each field, counted before filling. Labels is the number of permitted categorical labels. Filled $n$ counts non-empty answers among the 24,577 unique abstracts; it can include non-informative labels such as ``Not specified'' and excludes CDEs that could not be filled. The accompanying schema\_dictionary.json preserves the full questions, identifiers, and response vocabularies.
\par\smallskip
\noindent\textit{Interpretation of the generated schema.}
The questions and permissible value sets were generated by the AI pipeline and are reproduced without manual expansion or correction. They reflect the concepts and labels retained during schema induction, rather than a comprehensive clinical classification. Their inclusion does not establish that all relevant values were identified or correctly attributed.

\endgroup

\clearpage
\begingroup
\small
\setlength{\tabcolsep}{3pt}
\begin{longtable}{>{\raggedright\arraybackslash}p{4.6cm} >{\raggedright\arraybackslash}p{6.0cm} r r r}
\caption{Up to five most frequent exact labels per categorical field in 24,577 unique abstracts.}\label{supp:labels}\\
\toprule
Field & Exact label & Count & Field $n$ & \% \\
\midrule\endfirsthead
\toprule
Field & Exact label & Count & Field $n$ & \% \\
\midrule\endhead
\midrule\multicolumn{5}{r}{Continued on next page}\\\endfoot
\bottomrule\endlastfoot
15. Did Patient Develop Distant Metastases & No & 5,755 & 11,050 & 52.1 \\
 & Yes & 5,295 & 11,050 & 47.9 \\
16. Tumor Recurrence During Follow Up & Yes & 3,956 & 7,696 & 51.4 \\
 & No & 3,740 & 7,696 & 48.6 \\
17. Was Surgical Resection Performed & Yes & 13,397 & 14,933 & 89.7 \\
 & No & 1,536 & 14,933 & 10.3 \\
18. Granulocytic Sarcoma Anatomical Sites & Heart & 331 & 2,982 & 11.1 \\
 & Skin & 228 & 2,982 & 7.6 \\
 & Lungs & 190 & 2,982 & 6.4 \\
 & Breast & 189 & 2,982 & 6.3 \\
 & Soft tissues & 179 & 2,982 & 6.0 \\
19. Tumor Dimensions Cm & Large & 22 & 43 & 51.2 \\
 & 10 $\times$ 6 cm & 8 & 43 & 18.6 \\
 & $>$5 cm & 7 & 43 & 16.3 \\
 & 1 cm & 5 & 43 & 11.6 \\
 & 2 cm $\times$ 2 cm & 1 & 43 & 2.3 \\
20. Cytogenetic Abnormality Identified & EWSR1 gene rearrangement & 65 & 419 & 15.5 \\
 & SYT-SSX & 42 & 419 & 10.0 \\
 & BRAF V600E & 36 & 419 & 8.6 \\
 & ALK gene rearrangement & 24 & 419 & 5.7 \\
 & MDM2 amplification & 22 & 419 & 5.3 \\
21. Imaging Modality For Diagnosis & CT & 902 & 2,090 & 43.2 \\
 & MRI & 442 & 2,090 & 21.1 \\
 & CT scan & 300 & 2,090 & 14.4 \\
 & Transthoracic echocardiography & 102 & 2,090 & 4.9 \\
 & FDG-PET CT & 101 & 2,090 & 4.8 \\
22. Figo Stage & IA & 25 & 210 & 11.9 \\
 & IV & 24 & 210 & 11.4 \\
 & IB & 20 & 210 & 9.5 \\
 & IIIC & 19 & 210 & 9.0 \\
 & pT1 & 18 & 210 & 8.6 \\
23. Histologic Grade Of Tumor & High-grade malignant & 741 & 1,969 & 37.6 \\
 & Low & 462 & 1,969 & 23.5 \\
 & Grade II & 140 & 1,969 & 7.1 \\
 & Poorly differentiated & 124 & 1,969 & 6.3 \\
 & Malignant & 120 & 1,969 & 6.1 \\
24. Histological Classification Of Tumor & Sarcoma & 2,076 & 4,919 & 42.2 \\
 & Liposarcoma & 908 & 4,919 & 18.5 \\
 & Carcinosarcoma & 687 & 4,919 & 14.0 \\
 & Sarcomatoid Carcinoma & 659 & 4,919 & 13.4 \\
 & Fibrosarcoma & 376 & 4,919 & 7.6 \\
25. Histological Diagnosis Skin Tumor & Leiomyosarcoma & 1,004 & 4,717 & 21.3 \\
 & Kaposi sarcoma & 915 & 4,717 & 19.4 \\
 & Synovial sarcoma & 669 & 4,717 & 14.2 \\
 & Dermatofibrosarcoma protuberans & 321 & 4,717 & 6.8 \\
 & Malignant fibrous histiocytoma & 272 & 4,717 & 5.8 \\
26. Tumor Histological Pattern & leiomyosarcoma & 1,526 & 11,985 & 12.7 \\
 & Angiosarcoma & 1,480 & 11,985 & 12.3 \\
 & rhabdomyosarcoma & 764 & 11,985 & 6.4 \\
 & Liposarcoma & 743 & 11,985 & 6.2 \\
 & Kaposi sarcoma & 681 & 11,985 & 5.7 \\
27. Histological Subtype Of Tumor & Squamous Cell Carcinoma & 196 & 308 & 63.6 \\
 & Adenocarcinoma & 128 & 308 & 41.6 \\
28. Histological Type Of Tumor & Leiomyosarcoma & 2,038 & 13,117 & 15.5 \\
 & Spindle cell sarcoma & 1,392 & 13,117 & 10.6 \\
 & Ewing sarcoma & 1,197 & 13,117 & 9.1 \\
 & Osteogenic sarcoma & 1,065 & 13,117 & 8.1 \\
 & Cutaneous angiosarcoma & 1,005 & 13,117 & 7.7 \\
29. Histopathologic Diagnosis Biopsy & Sarcoma & 2,503 & 11,604 & 21.6 \\
 & Leiomyosarcoma & 1,634 & 11,604 & 14.1 \\
 & Angiosarcoma & 1,198 & 11,604 & 10.3 \\
 & Rhabdomyosarcoma & 983 & 11,604 & 8.5 \\
 & Ewing's sarcoma family of tumors & 814 & 11,604 & 7.0 \\
30. Patient Age At Diagnosis & Young Adults & 729 & 5,132 & 14.2 \\
 & Adolescent & 602 & 5,132 & 11.7 \\
 & 50-59 & 495 & 5,132 & 9.6 \\
 & 40-50 Years & 445 & 5,132 & 8.7 \\
 & 67-71 & 274 & 5,132 & 5.3 \\
31. Patient Race Or Ethnicity & Japanese & 232 & 1,057 & 21.9 \\
 & Caucasian & 187 & 1,057 & 17.7 \\
 & White & 175 & 1,057 & 16.6 \\
 & Chinese & 83 & 1,057 & 7.9 \\
 & African American & 69 & 1,057 & 6.5 \\
32. Patient Sex & Male & 10,690 & 19,806 & 54.0 \\
 & Female & 9,116 & 19,806 & 46.0 \\
33. Hiv Serostatus & Positive & 313 & 675 & 46.4 \\
 & AIDS & 162 & 675 & 24.0 \\
 & HIV-negative & 108 & 675 & 16.0 \\
 & Negative & 52 & 675 & 7.7 \\
 & Infected & 22 & 675 & 3.3 \\
34. Primary Hematologic Diagnosis & Kaposi Sarcoma & 697 & 1,755 & 39.7 \\
 & Myeloid Sarcoma & 376 & 1,755 & 21.4 \\
 & Acute Myeloid Leukemia & 154 & 1,755 & 8.8 \\
 & Non-Hodgkin's Lymphoma & 129 & 1,755 & 7.4 \\
 & Acute Myelogenous Leukemia & 61 & 1,755 & 3.5 \\
35. Metastatic Sites Reported & Lung & 799 & 3,084 & 25.9 \\
 & Bone & 415 & 3,084 & 13.5 \\
 & Liver & 395 & 3,084 & 12.8 \\
 & Pulmonary & 336 & 3,084 & 10.9 \\
 & Brain & 312 & 3,084 & 10.1 \\
36. Surgical Procedures Performed & Excision of tumor mass & 1,044 & 3,349 & 31.2 \\
 & Hysterectomy & 455 & 3,349 & 13.6 \\
 & Tumor excision & 392 & 3,349 & 11.7 \\
 & Bilateral salpingo‑ovarian oophorectomy & 316 & 3,349 & 9.4 \\
 & Wide excision of chest wall chondrosarcoma & 237 & 3,349 & 7.1 \\
37. Therapeutic Agent Administered & pazopanib & 99 & 892 & 11.1 \\
 & Methotrexate & 89 & 892 & 10.0 \\
 & pembrolizumab & 82 & 892 & 9.2 \\
 & Anlotinib & 71 & 892 & 8.0 \\
 & Nivolumab & 64 & 892 & 7.2 \\
38. Treatment Modalities Administered & Surgery & 6,211 & 8,402 & 73.9 \\
 & Chemotherapy & 2,523 & 8,402 & 30.0 \\
 & Radiotherapy & 1,663 & 8,402 & 19.8 \\
 & Surgical Excision & 586 & 8,402 & 7.0 \\
 & Corticosteroids & 198 & 8,402 & 2.4 \\
39. Primary Treatment Type & Surgery & 7,615 & 9,306 & 81.8 \\
 & Chemotherapy & 1,608 & 9,306 & 17.3 \\
 & Radiotherapy & 762 & 9,306 & 8.2 \\
 & Conservative surgery & 89 & 9,306 & 1.0 \\
 & Radical cystoprostatectomy & 34 & 9,306 & 0.4 \\
40. Surgical Procedure Type & Surgical excision & 1,177 & 4,877 & 24.1 \\
 & Surgical resection & 1,174 & 4,877 & 24.1 \\
 & Wide excision & 486 & 4,877 & 10.0 \\
 & Radical nephrectomy & 390 & 4,877 & 8.0 \\
 & Excision & 274 & 4,877 & 5.6 \\
41. Type Of Surgical Resection Primary Tumor & Excision & 898 & 3,471 & 25.9 \\
 & Complete Surgical Resection & 359 & 3,471 & 10.3 \\
 & Surgical Resection & 171 & 3,471 & 4.9 \\
 & Mastectomy & 120 & 3,471 & 3.5 \\
 & Wide Margin Excision & 112 & 3,471 & 3.2 \\
42. Primary Tumor Location 2 & Lung & 808 & 1,680 & 48.1 \\
 & Breast & 674 & 1,680 & 40.1 \\
 & Anterior left thigh & 198 & 1,680 & 11.8 \\
43. Disease Stage At Initial Diagnosis & Metastatic & 573 & 1,417 & 40.4 \\
 & Advanced & 256 & 1,417 & 18.1 \\
 & Localized & 111 & 1,417 & 7.8 \\
 & Stage IV & 93 & 1,417 & 6.6 \\
 & Stage III & 56 & 1,417 & 4.0 \\
44. Final Clinical Outcome & Alive & 8,788 & 12,266 & 71.6 \\
 & Dead & 3,478 & 12,266 & 28.4 \\
45. Surgical Margin Status & Negative & 1,168 & 1,365 & 85.6 \\
 & Positive & 181 & 1,365 & 13.3 \\
 & Close & 16 & 1,365 & 1.2 \\
46. Extramedullary Tumor Location & Retroperitoneal & 747 & 9,884 & 7.6 \\
 & Liver & 552 & 9,884 & 5.6 \\
 & Soft tissues & 451 & 9,884 & 4.6 \\
 & Bone & 407 & 9,884 & 4.1 \\
 & Pericardium & 402 & 9,884 & 4.1 \\
47. Kaposi Sarcoma Location & Scalp & 138 & 1,249 & 11.0 \\
 & Skin & 102 & 1,249 & 8.2 \\
 & Oral cavity & 72 & 1,249 & 5.8 \\
 & Pulmonary & 72 & 1,249 & 5.8 \\
 & Tongue & 59 & 1,249 & 4.7 \\
48. Metastatic Tumor Location & Lung & 659 & 5,397 & 12.2 \\
 & Liver & 497 & 5,397 & 9.2 \\
 & Brain & 357 & 5,397 & 6.6 \\
 & Lungs & 252 & 5,397 & 4.7 \\
 & Skin & 249 & 5,397 & 4.6 \\
49. Primary Lesion Location & Not specified & 1,561 & 10,938 & 14.3 \\
 & uterine body & 541 & 10,938 & 4.9 \\
 & lung & 507 & 10,938 & 4.6 \\
 & breast parenchyma & 495 & 10,938 & 4.5 \\
 & Retroperitoneal space surrounding duodenum & 366 & 10,938 & 3.3 \\
50. Primary Tumor Location & Uterus & 850 & 10,926 & 7.8 \\
 & Breast & 734 & 10,926 & 6.7 \\
 & Heart & 699 & 10,926 & 6.4 \\
 & Lung & 665 & 10,926 & 6.1 \\
 & Kidney & 643 & 10,926 & 5.9 \\
51. Tumor Anatomical Sites Involved & Liver & 670 & 1,421 & 47.1 \\
 & Skin & 589 & 1,421 & 41.4 \\
 & Spleen & 141 & 1,421 & 9.9 \\
 & Abdominal lymph nodes & 69 & 1,421 & 4.9 \\
 & Bilateral axillae & 24 & 1,421 & 1.7 \\
52. Chemotherapy Agents Used & Chemotherapy (unspecified) & 245 & 1,271 & 19.3 \\
 & Ifosfamide & 169 & 1,271 & 13.3 \\
 & doxorubicin & 154 & 1,271 & 12.1 \\
 & cisplatin & 121 & 1,271 & 9.5 \\
 & Vincristine & 117 & 1,271 & 9.2 \\
53. Preop Chemotherapy Agents & Ifosfamide & 205 & 721 & 28.4 \\
 & Vincristine & 192 & 721 & 26.6 \\
 & Doxorubicin & 192 & 721 & 26.6 \\
 & Cyclophosphamide & 146 & 721 & 20.2 \\
 & Cisplatin & 128 & 721 & 17.8 \\
\end{longtable}
\noindent Labels are not merged or recoded. Denominators are non-empty records for the individual field; multi-label percentages can sum above 100\%. Rankings break ties by first occurrence in the source export. An exact label can itself describe unspecified information or combine concepts.
\par\smallskip
\noindent This table reports up to five frequent extracted labels per field, rather than the complete permissible value sets. The underlying questions and vocabularies were generated by the AI pipeline and may be incomplete or ambiguous.
\endgroup